\documentclass[10pt,twocolumn,letterpaper]{article}

\usepackage[pagenumbers]{cvpr} 

\definecolor{hirow}{HTML}{F8D7DA}    
\definecolor{hicell}{HTML}{FFF3CD}   
\definecolor{best}{HTML}{FDEDEE}     

\newcommand{\bestnum}[1]{\cellcolor{best}\bfseries #1}
\newcommand{\second}[1]{\cellcolor{hicell}#1}

\usepackage{graphicx}
\usepackage{booktabs}

\usepackage{color}
\usepackage{bm}
\usepackage{multirow}
\usepackage{flushend}
\usepackage{caption}
\usepackage[table]{xcolor} 
\usepackage{amsmath, amssymb}

\usepackage{listings}
\usepackage{xcolor}

\usepackage{colortbl}
\usepackage{subcaption}

\usepackage{pifont}
\newcommand{\cmark}{\ding{51}}

\definecolor{cvprblue}{rgb}{0.21,0.49,0.74}
\usepackage[pagebackref,breaklinks,colorlinks,allcolors=cvprblue]{hyperref}

\def\paperID{6539} 
\def\confName{CVPR}
\def\confYear{2026}

\title{Lite Any Stereo: Efficient Zero-Shot Stereo Matching}

\author{
  Junpeng Jing \quad
  Weixun Luo \quad
  Ye Mao\footnotemark[2]  \quad
  Krystian Mikolajczyk \\
  Imperial College London \\
  {\tt\small \url{https://tomtomtommi.github.io/LiteAnyStereo/}}
}

\begin{document}
\renewcommand{\thefootnote}{\fnsymbol{footnote}}
\twocolumn[{
\renewcommand\twocolumn[1][]{#1}
\maketitle
\begin{center}
    \centering
    \vspace{-1.2em}
    \includegraphics[width=1\linewidth,page=1]{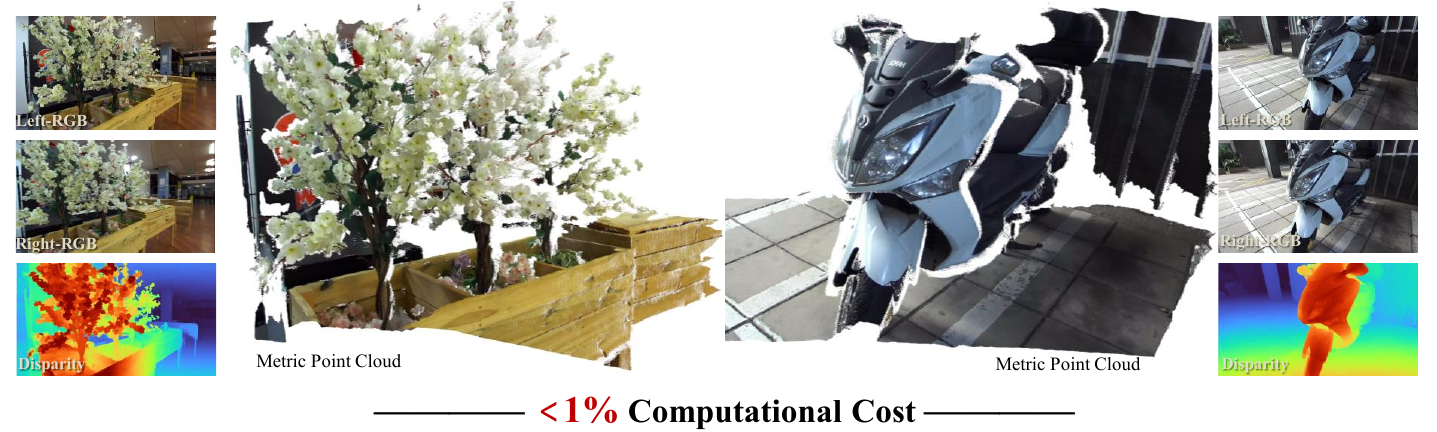}
    \vspace{-1.8em}
    \captionsetup{hypcap=false} 
    \captionof{figure}{Zero-shot predictions on in-the-wild stereo images. The proposed method produces accurate disparity estimates across diverse scenarios while maintaining high efficiency, requiring less than 1\% of the computational cost than accuracy-focused methods.}
    \label{fig:figure1}
\end{center}
}]

\footnotetext[2]{Corresponding author: \texttt{ye.mao21@imperial.ac.uk}}

\begin{abstract}
Recent advances in stereo matching have focused on accuracy, often at the cost of significantly increased model size. Traditionally, the community has regarded efficient models as incapable of zero-shot ability due to their limited capacity. In this paper, we introduce \textbf{Lite Any Stereo}, a stereo depth estimation framework that achieves strong zero-shot generalization while remaining highly efficient. To this end, we design a compact yet expressive backbone to ensure scalability, along with a carefully crafted hybrid cost aggregation module. We further propose a three-stage training strategy on million-scale data to effectively bridge the sim-to-real gap. Together, these components demonstrate that an ultra-light model can deliver strong generalization, ranking $\bm{1^{st}}$ across four widely used real-world benchmarks. Remarkably, our model attains accuracy comparable to or exceeding state-of-the-art non-prior-based accurate methods while requiring less than $\bm{1\%}$ computational cost, setting a new standard for efficient stereo matching.
\end{abstract}    
\section{Introduction}
\label{sec:intro}



From the foundational work of \cite{marr1988cooperative} through the classical advances culminating in works such as \cite{taniai2017continuous}, stereo vision has seen decades of steady progress built on a wide range of algorithmic ideas. In the past decade, deep learning methods have driven a dramatic leap in accuracy—leading many researchers to view stereo vision as largely solved—yet these gains often rely on large, computationally expensive models that are difficult to deploy on resource-limited hardware. 

Learning-based methods \cite{lipson2021raft, li2022practical, xu2023accurate, wang2024selective} achieve remarkable accuracy, continuously raising scores on standard benchmarks \cite{middlebury, eth3d, kitti12, kitti15}. These methods can be broadly categorized as accuracy-oriented. 
More recently, the emergence of foundation models trained on internet-scale data, such as the DepthAnything series \cite{yang2024depth, yang2024depth2}, has led to further advancement in the field. Stereo models augmented with depth priors \cite{wen2025foundationstereozeroshotstereomatching, cheng2025monstermarrymonodepthstereo, jiang2025defomstereodepthfoundationmodel} have demonstrated excellent zero-shot generalization: a single set of model weights can deliver strong results across diverse scenarios. However, despite these successes, such approaches prioritize accuracy over efficiency, which limits their practical applications.

In contrast, efficiency-oriented approaches~\cite{shamsafar2022mobilestereonet,guo2024lightstereochannelboostneed, xu2025banet} trade accuracy for faster inference and lower resource use, however, the accuracy gap to large models remains significant. 
The absence of light models with strong zero-shot generalization may lead to a conclusion that such models lack capacity for zero-shot applications. Consequently, most approaches still rely on domain-specific fine-tuning, thus falling short of being practical off-the-shelf solutions.

Training from large data is a key factor in enabling zero-shot capability.
StereoAnything \cite{guo2024stereo} scaled training by using a monocular depth model \cite{yang2024depth2} to generate pseudo-disparity maps, yielding 30M real-world samples. While large data improves zero-shot performance for efficient models \cite{guo2024lightstereochannelboostneed}, they remain far behind accuracy-oriented models, primarily due to the limited quality of monocular depth. 

\begin{figure}[t]
   \begin{center}
   \includegraphics[width=1\linewidth]{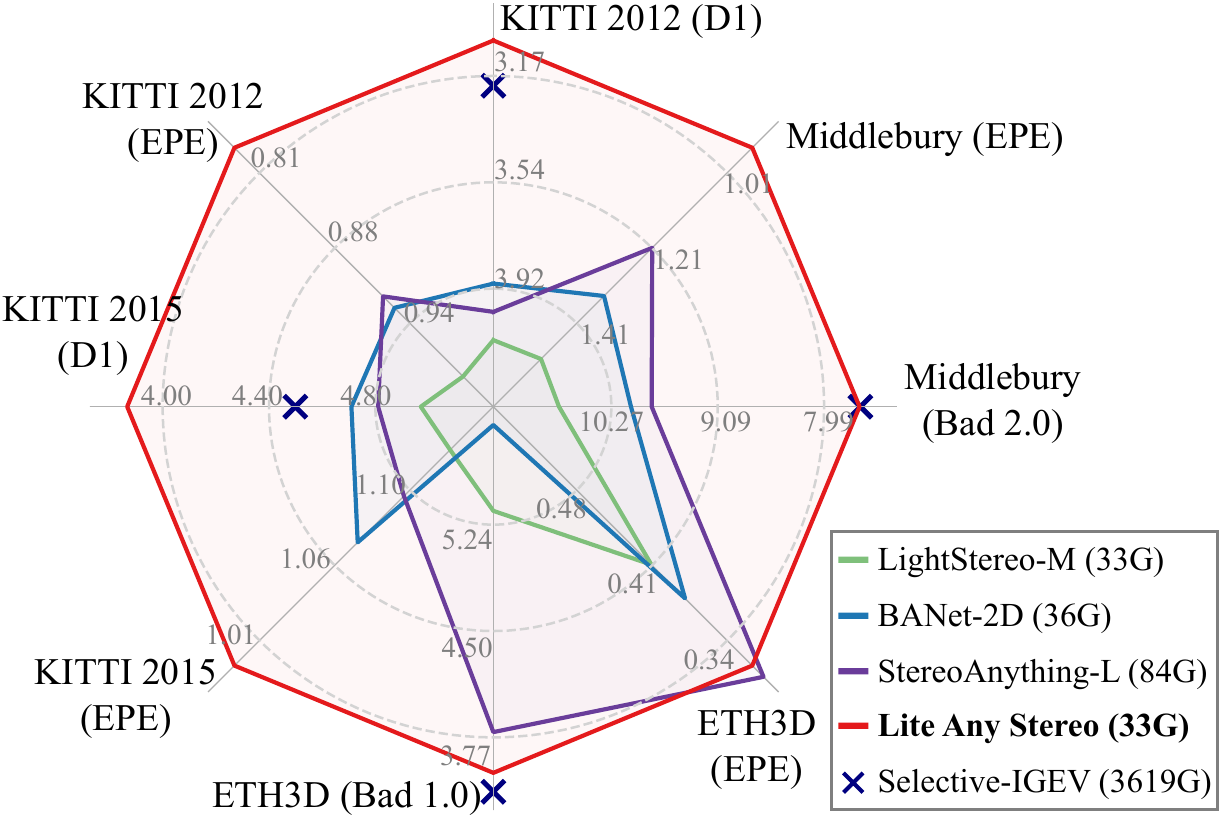}
   \end{center}
   \vspace{-.5em}
   \caption{Zero-shot performance. Our method achieves SOTA by a large margin, with even better or comparable non-prior-based accurate model, while requiring less than 1\% of their MACs.} 
   \vspace{-1.1em}
   \label{fig:figure2}
\end{figure}

In this paper, we propose Lite Any Stereo, an ultra-lite stereo matching model designed for zero-shot generalization. We achieve this from two perspectives: architecture and training strategy. We design a backbone with a hybrid cost aggregation module that jointly leverages 2D and 3D representations, capturing complementary spatial and disparity cues. This carefully crafted backbone achieves very low computational cost and fast inference speed. To train for zero-shot capability, we scale the data to a million level with a three-stage strategy. Specifically, self-distillation on synthetic labeled data is performed after supervised training. We further make use of real-world unlabeled data, which has long been overlooked in stereo, for knowledge distillation. This combination effectively mitigates the sim-to-real gap and proves to be universally beneficial across varied model architectures. Together, these innovations significantly improve performance. As shown in Fig.~\ref{fig:figure1}, our method shows strong generalization, producing accurate disparities on in-the-wild images with high efficiency.

{Our main contributions can be summarized as follows:}
\begin{itemize}
\item We present {Lite Any Stereo}, an efficient stereo matching model that achieves zero-shot capability. As shown in Fig.~\ref{fig:figure2}, it delivers the highest zero-shot accuracy among all efficient methods by a large margin. This is the first efficient model to outperform or match accuracy-oriented models that do not use foundational priors, while requiring less than 1\% of their computational cost.
\item We propose a hybrid cost aggregation module that captures complementary disparity and spatial information, enhancing representation at low cost.
\item We develop a three-stage training strategy that integrates synthetic supervision, self-distillation, and real-world knowledge distillation, effectively improving zero-shot generalization across diverse architectures.
\end{itemize}

\section{Related Work}

In this section, we first review accurate stereo matching methods, followed by a discussion of approaches focused on computational efficiency.


\subsection{Accurate Stereo Methods}
Modern stereo matching methods based on deep learning typically construct either 3D or 4D cost volumes to capture pixel-level correspondences between the left and right views. These volumes serve as the foundation for disparity estimation, with convolutional neural networks used to aggregate contextual information. Broadly, CNN-based approaches can be categorized into two types: those that emphasize cost aggregation~\cite{pang2017cascade, chang2018pyramid, liang2018learning, yang2019hierarchical, zhang2019ga, guo2019group, xu2020aanet, cheng2020hierarchical, gu2020cascade, tankovich2021hitnet, guo2023openstereo, guo2025stereocarla}, and those that adopt iterative refinement strategies~\cite{lipson2021raft, li2022practical, xu2023iterative, wang2024selective} to progressively enhance disparity predictions. Another line of research explores transformer-based stereo models~\cite{guo2022context, li2021revisiting, su2022chitransformer, weinzaepfel2023croco, xu2023unifying}, which leverage attention mechanisms to capture global context. In addition, some methods explicitly address temporal consistency in stereo videos~\cite{karaev2023dynamicstereo, jing2024matchstereovideos, jing2024matchstereovideosbidirectional, jing2025stereovideotemporallyconsistent}, aiming to produce consistent disparities across time. While these models perform well on domain-specific benchmarks~\cite{kitti12, kitti15, middlebury, eth3d}, generalization to unseen domains in zero-shot settings remains a major challenge. To address this, some works have focused on learning domain-invariant representations \cite{chuah2022itsa, zhang2022revisiting, chang2023domain, rao2023masked} and robust modules~\cite{shen2021cfnet, Jing_2023_ICCV, guo2024stereo}, aiming to improve performance under domain shifts. A recent new direction involves leveraging prior knowledge from monocular foundation models~\cite{yang2024depth, yang2024depth2}. By integrating monocular depth cues into stereo pipelines, several methods~\cite{bartolomei2024stereo, zhang2024learning, zhou2024all, cheng2025monstermarrymonodepthstereo, jiang2025defomstereodepthfoundationmodel, wen2025foundationstereozeroshotstereomatching} demonstrate enhanced zero-shot generalization. However, these improvements often come at the cost of increased computational complexity, limiting their deployment in real-time or resource-constrained scenarios.

\begin{figure*}[t]
   \begin{center}
   \includegraphics[width=1\linewidth]{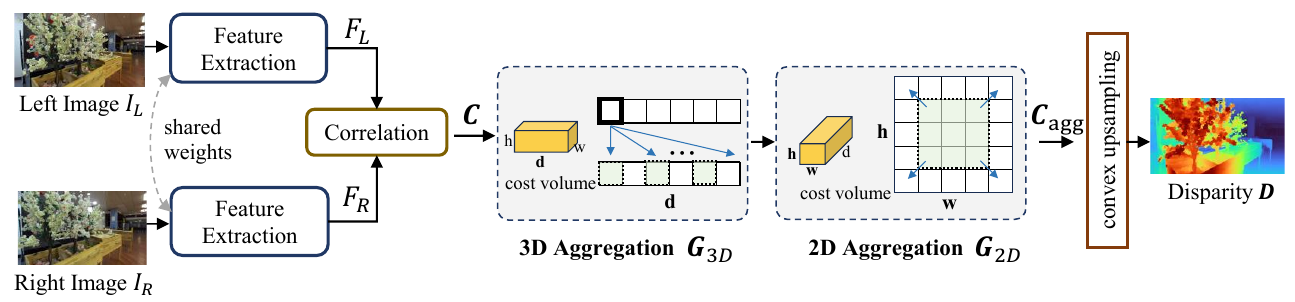}
   \end{center}
   \vspace{-1.5em}
    \caption{Overview of the proposed {Lite Any Stereo}. Given an input stereo image pair, features are first extracted using a shared-weight feature extraction module. A correlation module then constructs cost volume from extracted features, which is processed by a hybrid 3D-2D cost aggregation module to obtain aggregated cost volume along both disparity and spatial dimensions. Finally, low-resolution disparity map is estimated and a convex upsampling operation is applied to recover the full-resolution disparity map.}
    \vspace{-.8em}
   \label{fig:framework}
\end{figure*}

\begin{figure*}[t]
   \begin{center}
   \includegraphics[width=1\linewidth]{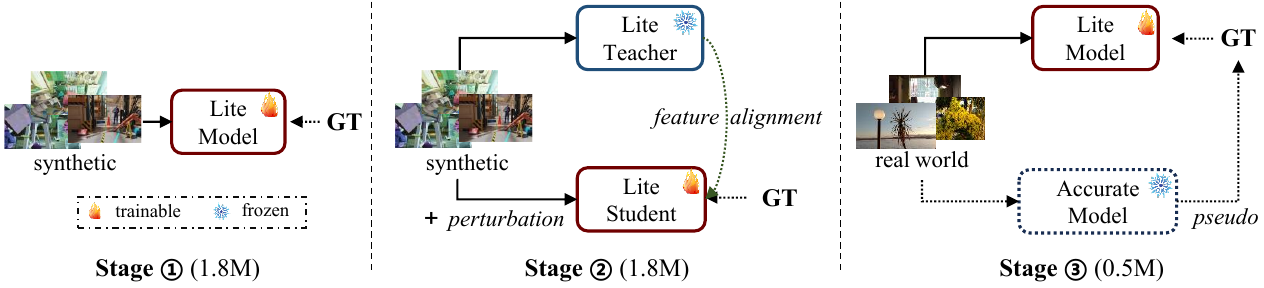}
   \end{center}
   \vspace{-1.2em}
    \caption{Overview of the proposed three-stage training strategy. \textbf{Stage \ding{172}:} The lite model is trained using a standard supervised setup on a mixed of synthetic datasets including 1.8M labeled stereo image pairs. \textbf{Stage \ding{173}:} Self-distillation is employed, where both teacher and student models are initialized from the Stage \ding{172} weights. The teacher receives clean data, while the student is fed perturbed inputs to encourage learning of domain-invariant representations via feature alignment. \textbf{Stage \ding{174}:} The lite model is fine-tuned on unlabeled real-world data using pseudo labels generated by a frozen accurate model.}
    \vspace{-.6em}
   \label{fig:training_strategy}
\end{figure*}

\subsection{Efficient Stereo Matching}
Real-time performance is critical for stereo matching in practical applications. Early approaches~\cite{khamis2018stereonet, duggal2019deeppruner, wang2020fadnet} address this by estimating disparities at reduced spatial resolutions to lower computational cost, though at the expense of accuracy. Subsequent methods aim to improve this efficiency and accuracy trade-off based on 3D aggregation. Representative advancements include guided cost volume excitation in CoEx~\cite{bangunharcana2021correlate}, edge-aware upsampling in BGNet~\cite{xu2021bilateral}, and sparse attention for selective high-resolution matching in Fast-ACVNet~\cite{xu2022acvnet, xu2023accurate}. To further reduce complexity, another line of works have explored 2D-based alternatives to traditional 3D convolutions. Notable examples include AANet~\cite{xu2020aanet}, which employs deformable 2D convolutions for adaptive cost aggregation; HITNet~\cite{tankovich2021hitnet}, which reconstructs disparities through iterative warping without explicit cost volumes; and MobileStereoNet-2D~\cite{shamsafar2022mobilestereonet}, which adopts lightweight MobileNet blocks. More recent efforts continue this trend with specialized strategies: uncertainty-aware adaptive warping in Lite-CREStereo++~\cite{Jing_2023_ICCV}, channel-wise enhancement in LightStereo~\cite{guo2024lightstereochannelboostneed}, and frequency-disparity guided bilateral aggregation in BANet~\cite{xu2025banet}. Despite these advances in efficiency, most of these models are optimized for a specific domain, particularly the KITTI online benchmarks~\cite{kitti12, kitti15}, and still struggle to generalize well across diverse, unseen scenarios. Achieving strong zero-shot ability while maintaining real-time efficiency remains an open challenge.

\section{Method}

In this section, we introduce Lite Any Stereo, an efficient feed-forward network for zero-shot stereo matching. We first demonstrate the overall framework (Sec. \ref{sec: framework}), and then go through the training strategy (Sec. \ref{sec: training strategy}).

\subsection{Framework} \label{sec: framework}
As shown in Fig.~\ref{fig:framework}, the overall framework comprises four main stages: feature extraction, correlation, cost aggregation, and disparity estimation.

\noindent \textbf{Feature Extraction.} Recent methods~\cite{jiang2025defomstereodepthfoundationmodel, cheng2025monstermarrymonodepthstereo, wen2025foundationstereozeroshotstereomatching} have demonstrated remarkable improvements by leveraging depth features from DepthAnything (DA)~\cite{yang2024depth, yang2024depth2}. However, we find that the additional computational overhead introduced by DA, even with the smallest DA-S variant, is prohibitive for an efficiency-oriented stereo model. Therefore, we adopt a conventional backbone \cite{guo2024lightstereochannelboostneed,xu2025banet} pretrained on ImageNet \cite{deng2009imagenet} without external priors for feature extraction. Moreover, we find that compared with more recent CNN  backbones~\cite{liu2022convnet, woo2023convnext}, the channel configuration of MobileNetV2~\cite{sandler2018mobilenetv2} is better aligned with the requirements of our task. Specifically, given a pair of rectified stereo images $\{\mathbf{I}_{L}, \mathbf{I}_{R}\} \in \mathbb{R}^{H\times W\times 3}$, we employ two weight-sharing networks to pyramidally generate multi-scale features $\{\mathbf{F}_{L}^s\}$ and $\{\mathbf{F}_{R}^s\}$, where $s \in \left\{\frac{1}{4}, \frac{1}{8}, \frac{1}{16}, \frac{1}{32}\right\}$ denotes the downsampling ratio. To unify the spatial resolution for subsequent processing, features at all scales are upsampled to $\tfrac{1}{4}$ resolution using residual upsampling blocks, following~\cite{guo2024lightstereochannelboostneed}.


\noindent \textbf{Correlation.} Given the extracted left and right feature maps $\mathbf{F}_{L}^{\frac{1}{4}}$ and $\mathbf{F}_{R}^{\frac{1}{4}}$, the cost volume $\mathbf{C}$ for each disparity $d$ within the range $[0, D_{\mathrm{max}} / 4]$ is constructed as:  
\begin{equation}
    \mathbf{C}(d, h, w) = \frac{1}{N_c} \left\langle \mathbf{F}_{L}^{\frac{1}{4}}(h, w), \ \mathbf{F}_{R}^{\frac{1}{4}}(h, w-d) \right\rangle ,
    \label{eq:correlation}
\end{equation}
where $D_{\max}$ denotes the predefined maximum disparity value, $\langle \cdot, \cdot \rangle$ denotes the inner product, $N_c$ is the number of channels, and $(h,w)$ represents the pixel location.

\begin{figure}[t]
   \begin{center}
   \includegraphics[width=1\linewidth]{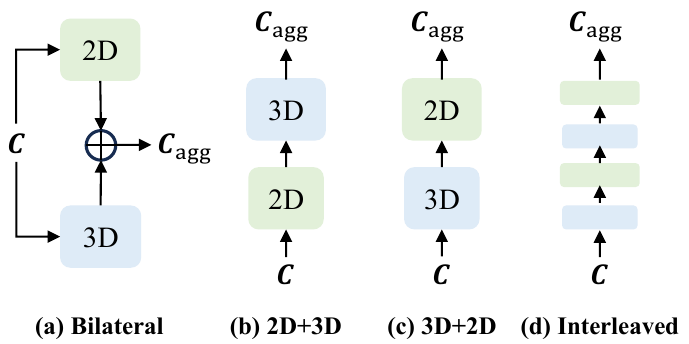}
   \end{center}
   \vspace{-1.5em}
    \caption{Design choices for hybrid cost aggregation module.} 
    \vspace{-.6em}
    \label{fig:backbone_comparison}
\end{figure}


\noindent \textbf{Cost Aggregation.} Recent efficient models adopt 2D cost aggregation~\cite{shamsafar2022mobilestereonet,guo2024lightstereochannelboostneed,xu2025banet}, which reduces computation by collapsing the disparity into channels and performing spatial-only convolutions. However, this strategy does not explicitly model structured continuity along the disparity dimension, potentially limiting generalization. To address this, we introduce a hybrid aggregation module that combines complementary 2D and 3D reasoning. The 3D block $\mathbf{G}_{\text{3D}}(\cdot)$ jointly aggregates over disparity and spatial dimensions to capture cross-disparity structure, while the 2D block $\mathbf{G}_{\text{2D}}(\cdot)$ provides efficient spatial refinement. This design preserves geometric awareness with minimal overhead.

Specifically, multi-scale 3D convolutions are employed in the 3D block, and ConvNeXt layers~\cite{liu2022convnet} are used in the 2D block. A naive half–half hybrid design is ineffective, as 3D convolutions dominate the compute budget along the disparity dimension, even though many disparity levels contribute little to the final results. Instead, we retain only a small proportion of 3D component to preserve disparity perception while maintaining efficiency. We also explore multiple design choices for integrating 2D and 3D branches while keeping the overall computational cost fixed, as illustrated in Fig.~\ref{fig:backbone_comparison}. In particular:  (a) follows~\cite{xu2025banet}, where bilateral aggregation is adopted and the outputs of 2D and 3D branches are summed;  (b) and (c) apply serial connections, but with different ordering of 2D and 3D blocks;  (d) interleaves 2D and 3D blocks. Through ablation studies (Section~\ref{sec:Ablation Study}), we find that design (c) achieves the best performance. Therefore, we adopt (c) as the default configuration of our cost aggregation module, formulated as follows,
\begin{equation}
\mathbf{C}_{\text{agg}} = \mathbf{G}_{\text{2D}}(\, \mathbf{G}_{\text{3D}}(\mathbf{C}) \,).
\end{equation}

\noindent \textbf{Disparity Estimation.} Similar to other efficient methods \cite{guo2024lightstereochannelboostneed, xu2025banet}, we apply the soft-argmax operation to regress the disparity map $\mathbf{d}$ at $\frac{1}{4}$ scale:
\begin{equation}
    \mathbf{d} = \sum_{d=0}^{D_{\max}/4 - 1} d \times \sigma(\mathbf{C}_{\text{agg}}(d)),
    \label{eq:disp0}
\end{equation}
where $\sigma(\cdot)$ is a softmax layer. Convex upsampling is then used to upsample $\mathbf{d}$ to the full-resolution $\mathbf{D} \in \mathbb{R}^{H \times W}$.

\begin{table}[t] \addtolength{\tabcolsep}{-4pt}
\caption{Overview of the real-world stereo datasets (0.5M samples in total) used for training our model.}
\vspace{-.6em}
\centering
\small
\begin{tabular}{lccccr}
\toprule
\textbf{Dataset} & \textbf{Indoor} & \textbf{Outdoor} & \textbf{MPix} & \textbf{Images} \\
\midrule
Flickr1024~\cite{Flickr1024} & \cmark & \cmark  & 0.73 &  1K \\
InStereo2k~\cite{bao2020instereo2k} & \cmark &  & 0.93 & 2K \\
Holopix50K~\cite{hua2020holopix50k} & \cmark & \cmark  & 0.74 &  49K \\
Driving Stereo~\cite{yang2019drivingstereo} &  & \cmark  & 0.40 &  174K \\
SouthKenSV~\cite{jing2024matchstereovideosbidirectional} & \cmark & \cmark & 0.92 &  113K \\
UASOL \cite{bauer2019uasol} &  & \cmark & 2.74 &   156K \\
\bottomrule
\end{tabular}
\label{tab:dataset_overview}
\end{table}

\begin{figure}[t]
   \begin{center}
   \includegraphics[width=1\linewidth]{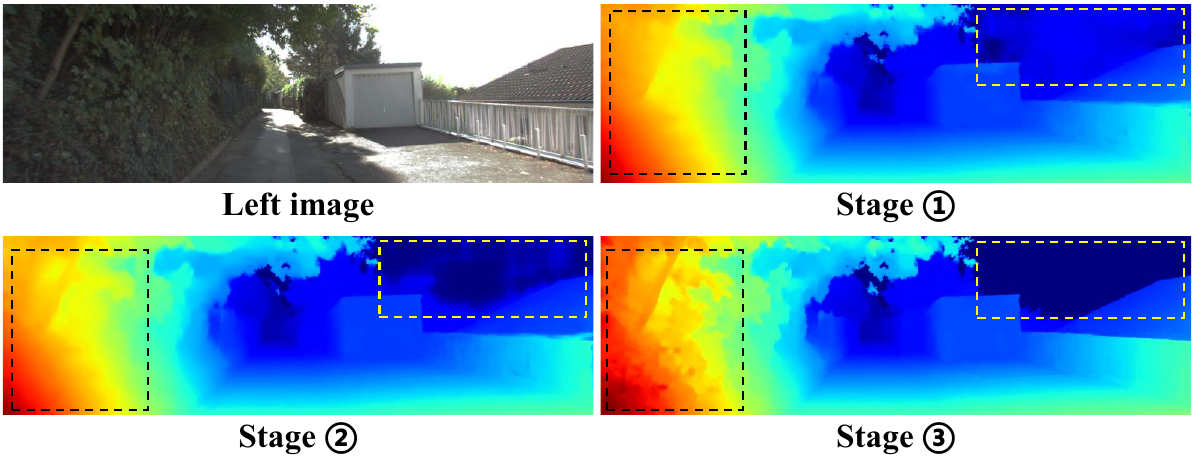}
   \end{center}
   \vspace{-1.5em}
    \caption{Effects of the proposed three stages training strategy.} 
    \vspace{-.8em}
    \label{fig:ablation_compare}
\end{figure}

\begin{table*}[t]
\caption{Ablation study on KITTI 2012 \cite{kitti12} (K.12), KITTI 2015 \cite{kitti15} (K.15), ETH3D \cite{eth3d} (E.), and Middlebury \cite{middlebury} (M.). We report D1 for KITTI, Bad 1 for ETH3D, and Bad 2 for Middlebury. We also report MACs (Multiply-Accumulate Operations) for backbone design choices. Unless otherwise specified, models are trained for 200K iterations without data augmentation on a 1.4M-image subset from synthetic datasets \cite{wen2025foundationstereozeroshotstereomatching, li2022practical, cabon2020vkitti2, fallingthings, mayer2016large} using the default operations in \cite{xu2025banet}. Default settings in final model are marked in \colorbox{gray!20}{gray}.}
\vspace{-.3em}

\subfloat[{Design choices for cost aggregation module, as shown in Fig. \ref{fig:backbone_comparison}. Introducing 3D block before 2D block in serial connection is effective.}]{
\begin{minipage}{0.46\linewidth}
\centering
\small
\begin{tabular}{c|cccc|c}
case & K.12 & K.15 & E. & M. & MACs (G)\\
\hline
2D  & 5.02 & 5.01 & 6.48 & 11.29 &  \textbf{32.9}  \\
bilateral  &  5.10 & 5.10 & 8.55 & 12.00 & 35.8 \\
2D-3D   & 4.73 & \textbf{4.44} & 28.85 & 11.52 & 33.9 \\
3D-2D & \cellcolor{gray!20}{4.78} & \cellcolor{gray!20}{4.64} & \cellcolor{gray!20}\textbf{5.39} & \cellcolor{gray!20}\textbf{10.89}  & \cellcolor{gray!20}{33.9}\\
interleaved & \textbf{4.61} & 4.73 & 6.20 & 11.34 & 35.2 \\

\end{tabular}
\end{minipage}
}
\hfill
\subfloat[{3D kernel size for cost aggregation module.} Enlarging the kernel size along disparity dimension is not effective. We keep the general $(3,3,3)$.]{
\begin{minipage}{0.46\linewidth}
\centering
\small
\begin{tabular}{c|cccc|c}
3D kernel & K.12 & K.15 & E. & M. & MACs (G)  \\
\hline
(3, 3, 3)  & \cellcolor{gray!20}\textbf{4.38} & \cellcolor{gray!20}\textbf{4.84} & \cellcolor{gray!20}\textbf{5.75} & \cellcolor{gray!20}\textbf{9.50}  & \cellcolor{gray!20}{32.8}\\
(5, 3, 3)   & 4.65 & 4.88 & 6.29 & 10.36 & 32.5 \\
(7, 3, 3)   & 4.84 & 4.85 & 6.37 & 9.64 & 32.6 \\
(11, 3, 3)  & 4.86 & 4.87 & 6.87 & 10.39 & 32.8 \\
(3, 1, 1)   & 4.70 & 4.85 & 6.63 & 10.06 & \textbf{32.1} \\
\end{tabular}
\end{minipage}
}

\vspace{.6em}
\subfloat[{Layer choices for 2D block in cost aggregation module. ConvNext is effective with a lowest computation cost.}]{
\begin{minipage}{0.46\linewidth}
\centering
\small
\begin{tabular}{c|cccc|c}
case & K.12 & K.15 & E. & M. & MACs (G) \\
\hline
MobileNet v2 & 4.78 & \textbf{4.64} & 5.39 & 10.89 & 33.9 \\
ConvNeXt & \cellcolor{gray!20}\textbf{4.38} & \cellcolor{gray!20}{4.84} & \cellcolor{gray!20}{5.75} & \cellcolor{gray!20}\textbf{9.50}  & \cellcolor{gray!20}\textbf{32.8}\\
ConvNeXt v2 & 4.79 & 4.81 & \textbf{5.03} & 10.52 & 34.0 \\
\end{tabular}
\end{minipage}
}
\hfill
\subfloat[{3D proportion for cost aggregation module.} 4.8\% is effective, while a larger one can be detrimental under the same computational budget.]{
\begin{minipage}{0.46\linewidth}
\centering
\small
\begin{tabular}{c|cccc|c}
3D proportion & K.12 & K.15 & E. & M. & MACs (G) \\
\hline
4.8\%   & \cellcolor{gray!20}\textbf{4.38} & \cellcolor{gray!20}{4.84} & \cellcolor{gray!20}{5.75} & \cellcolor{gray!20}\textbf{9.50}  & \cellcolor{gray!20}{32.8}\\
9.5\%   & 4.71 & \textbf{4.51} & 5.81 & 10.06 & \textbf{32.6} \\
15.6\%   & 4.49 & 4.68 & \textbf{5.54} & 10.34 & 32.7 \\
\end{tabular}
\end{minipage}
}

\vspace{.6em}
\subfloat[Training strategy. knowledge distillation is more effective than direct data augmentation.]{
\begin{minipage}{0.3\linewidth}
\small
\raggedright
\setlength{\tabcolsep}{4.5pt}
\begin{tabular}{c|cccc}
case & K.12 & K.15 & E. & M.  \\
\hline
none  & {4.38} & {4.84} & {5.75} & {9.50}\\
data aug. & 4.31 & 4.79 & \textbf{5.73} & 11.14  \\
know. dis. & \cellcolor{gray!20}\textbf{3.64} & \cellcolor{gray!20}\textbf{4.63} & \cellcolor{gray!20}{6.82} & \cellcolor{gray!20}\textbf{8.86}\\
\end{tabular}
\end{minipage}
}
\hfill
\subfloat[Knowledge distillation choices in stage \ding{173}. Fixed teacher's weights is more effective.]{
\begin{minipage}{0.3\linewidth}
\small
\raggedright
\setlength{\tabcolsep}{4.3pt}
\begin{tabular}{c|cccc}
case & K.12 & K.15 & E. & M.  \\
\hline
EMA & 3.97 & 4.82 & 6.91 & 9.78  \\
hard copy  & 4.22 & 4.71 & \textbf{6.35} & 10.07 \\
fixed & \cellcolor{gray!20}\textbf{3.64} & \cellcolor{gray!20}\textbf{4.63} & \cellcolor{gray!20}{6.82} & \cellcolor{gray!20}\textbf{8.86}\\
\end{tabular}
\end{minipage}
}
\hfill
\subfloat[Effects of the three stages training strategy on full training set. See Fig. \ref{fig:ablation_compare} for visualization.]{
\begin{minipage}{0.3\linewidth}
\small
\raggedright
\setlength{\tabcolsep}{5.5pt}
\begin{tabular}{c|cccc}
case & K.12 & K.15 & E. & M.  \\
\hline
stage  \ding{172}  & 4.05 & 4.55 & 4.43 & 8.49 \\
stage  \ding{173}  & 3.66 & 4.53 & 4.69 & \textbf{7.03}  \\
stage  \ding{174}  & \cellcolor{gray!20}\textbf{3.04} & \cellcolor{gray!20}\textbf{3.87} & \cellcolor{gray!20}\textbf{3.53} & \cellcolor{gray!20}{7.51}\\
\end{tabular}
\end{minipage}
}
\vspace{-.8em}
\label{tab:mae_ablation}
\end{table*}

\subsection{Training Strategy} \label{sec: training strategy}
To achieve zero-shot capability, the proposed model is trained on a large amount of high-quality data by following carefully designed training stages, which we illustrate in Fig.~\ref{fig:training_strategy}. We compile datasets into two categories: synthetic annotated data and realistic unannotated data. Since synthetic datasets provide accurate ground-truth annotations, we first train our model on them from scratch to enhance model's matching ability. Our synthetic datasets combine SceneFlow \cite{mayer2016large} (35K), FallingThings \cite{fallingthings} (30K), FSD \cite{wen2025foundationstereozeroshotstereomatching} (1.1M), CREStereo \cite{li2022practical} (0.2M), VKITTI2 \cite{cabon2020vkitti2} (21K), TartanAir \cite{tartanair2020iros} (0.31M), and Dynamic Replica \cite{karaev2023dynamicstereo} (0.14M), resulting in a total of 1.8M annotated samples. While other synthetic datasets are available, such as IRS \cite{wang2019irs}, Sintel \cite{sintel}, Spring \cite{Mehl2023_Spring}, and InfinigenSV \cite{jing2024matchstereovideosbidirectional}, they are excluded due to annotation quality issues (e.g., inaccurate disparities for transparent objects in IRS) or domain mismatch (e.g., cinematic or natural scenes in Spring and InfinigenSV). 

In \textbf{Stage \ding{172}}, the model is trained in a supervised, end-to-end manner without data augmentation, using the commonly used disparity loss $\mathcal{L}_{disp}$:
\begin{equation}
    \mathcal{L}_{disp} =  smooth_{L_1} (\mathbf{D} - \mathbf{D}_{gt}),
    \label{eq:loss}
\end{equation}
where $\mathbf{D}_{gt}$ denotes the ground-truth disparity.

In \textbf{Stage \ding{173}}, we introduce a self-distillation strategy to improve feature robustness. Both teacher and student models have the same architectures initialized from the first stage. The teacher model receives clean inputs, while the student model is exposed to strongly perturbed inputs (see supplementary materials), encouraging domain-invariant representation learning. We employ a feature alignment loss $\mathcal{L}_{feat}$ in addition to $\mathcal{L}_{disp}$:
\begin{equation}
    \mathcal{L}_{feat} = 1 - \frac{1}{HW} \sum_{i=1}^{HW} \cos(F_i, F'_i),
    \label{eq:feat_loss}
\end{equation}
where $F_i$ and $F'_i$ are feature vectors from the teacher and student models, respectively. Here, we evaluate several distillation schemes: (a) training only the student model with fixed teacher model weights, (b) updating the teacher model via Exponential Moving Average (EMA) \cite{polyak1992acceleration}, and (c) directly copying student model weights to the teacher model at each iteration. Our ablation study (Section~\ref{sec:Ablation Study}) shows that the simplest strategy (a) yields the best performance, and is therefore adopted in the second stage.

High-quality real-world stereo data with annotations remain scarce and sparse (e.g., Lidar-based ground truth), which constrains training scalability. In contrast, the vast amount of unlabeled real-world stereo data has not been used by the community. With the increasing availability and diversity of such data, in \textbf{Stage \ding{174}}, we leverage unannotated stereo pairs to improve model generalization, collecting a total of 0.5M pairs, as summarized in Tab.~\ref{tab:dataset_overview}. Pseudo labels are generated using FoundationStereo \cite{wen2025foundationstereozeroshotstereomatching}, a strong accurate model. Importantly, even for datasets with ground-truth annotations (e.g., DrivingStereo \cite{yang2019drivingstereo}), we use pseudo but {dense} labels for stereo pairs (excluding the weather subset), as they provide stronger supervision than sparse data.

We further observe that data quality is more critical than scale: incorporating low-quality or domain-specific data in the third stage can degrade zero-shot performance. For example, Stereo4D \cite{jin2024stereo4d} contains 18M stereo pairs mined from internet videos but only at a limited resolution (512$\times$512), HRWSI \cite{hrwsi} suffers from poor rectification quality, and datasets such as SCOD \cite{scod} are restricted to narrow domains. Finally, we do not apply self-distillation here, as it provides no observable performance gains when training with pseudo labels. As shown in Fig.~\ref{fig:ablation_compare}, the proposed strategy progressively improves the disparity estimation, making the leaves and background regions noticeably clearer.

\section{Experiments}

\begin{table*}[t] \addtolength{\tabcolsep}{4pt}
\caption{Zero-shot generalization results on four public benchmarks: KITTI 2012 \cite{kitti12}, KITTI 2015 \cite{kitti15}, ETH3D \cite{eth3d}, and Middlebury (H) \cite{middlebury}. The most commonly used metrics are adopted. In the first block, all efficient methods are trained only on Scene Flow \cite{mayer2016large}. In the second and third blocks, methods are allowed to train on any existing datasets excluding the four target domains. Accuracy-based methods are shown as reference. The weights and parameters are fixed for evaluation. MACs (G) are measured at the KITTI resolution of $1242 \times 375$. $^{*}$ indicates models retrained with original code on the same synthetic data as ours. $^{\dag}$ denotes results trained on 30M pseudo-labeled samples using the strategy in \cite{guo2024stereo}. $^{\ddag}$ marks results reported by \cite{wen2025foundationstereozeroshotstereomatching}. The \colorbox{best}{\textbf{best}} and \colorbox{hicell}{second best} are marked with colors.}
\vspace{-.6em}
\centering
\small
\begin{tabular}{lccccccccc} 
\toprule
\multirow{2}{*}{Method} &  \multicolumn{2}{c}{KITTI 2012} & \multicolumn{2}{c}{KITTI 2015} & \multicolumn{2}{c}{ETH3D} & \multicolumn{2}{c}{Middlebury} & \multirow{2}{*}{MACs (G)}\\ 
& D1 & EPE & D1  & EPE & Bad 1.0 & EPE & Bad 2.0 & EPE\\ 
\hline
\rowcolor{gray!10} \multicolumn{10}{l}{\hspace{0pt}\textit{Efficient methods: SceneFlow}}\\
CoEX \cite{bangunharcana2021correlate}  & 22.30 & 3.37 & 17.33 & 3.00 & 31.97 & 14.05 & 26.42 & 4.90 & 54\\
MobileStereoNet-2D \cite{shamsafar2022mobilestereonet} & 19.30 & 2.51 & 21.88 & 2.85 & 17.10 & 1.83 & 37.98 & 7.54 & 127 \\
MobileStereoNet-3D \cite{shamsafar2022mobilestereonet} & 19.59 & 2.67 & 17.89 & 3.01 & 18.42 & 3.02 & 25.26 & 4.61 & 564\\
FastACV \cite{xu2022acvnet} & 13.90 & 2.05 & 11.83 & 2.21 & 7.84 & 1.05 & 19.61 & 4.66 & 72\\
FastACV+ \cite{xu2023accurate} & 16.69 & 2.29 & 15.28 & 3.27 & 11.50 & 2.43 & 27.34 & 7.16 & 85\\
Lite-CREStereo++ \cite{Jing_2023_ICCV} & \second{5.93} & 1.29 & 7.37 & 1.36 & \bestnum{8.95} & {1.18} & \second{14.91} & 3.32 & 101 \\
LightStereo-M \cite{guo2024lightstereochannelboostneed} & 6.76 & \second{1.26} & 6.79 & 1.42 & 13.93 & \second{0.73} & 16.99 & \second{2.06} & \bestnum{33}\\
LightStereo-L \cite{guo2024lightstereochannelboostneed} & 6.80 & 1.27 & \second{6.62} & \bestnum{1.29} & \second{9.66} & \bestnum{0.50} & 17.23 & 2.88 & 84 \\
BANet-2D \cite{xu2025banet} & 14.75 & 2.23 & 16.98 & 3.91 & 44.89 & 35.95 & 26.79 & 6.96 & 36 \\
BANet-3D \cite{xu2025banet} & 17.39 & 2.41 & 17.28 & 3.36 & 29.27 & 14.06 & 28.79 & 8.05 & 78\\
\textbf{Lite Any Stereo} & \bestnum{5.45} & \bestnum{1.18} & \bestnum{6.45} & \second{1.32} & {15.38} & {0.77} & \bestnum{13.13} & \bestnum{1.60} &  \bestnum{33} \\
\hline
\rowcolor{gray!10} \multicolumn{10}{l}
{\hspace{0pt}\textit{Efficient methods: Million-scale}}\\
LightStereo-M$^{*}$ \cite{guo2024lightstereochannelboostneed} & 4.10 & 0.99 & 4.97 & 1.13 & 5.33 & 0.41 & 10.85 & 1.51 &  \bestnum{33} \\
BANet-2D$^{*}$ \cite{xu2025banet} & \second{3.90} & 0.93 & \second{4.71} & \second{1.07} & 5.92 & 0.38 & 10.05 & 1.34 &  36 \\
StereoAnything-L$^{\dag}$ \cite{guo2024stereo} & 4.00 & \second{0.92} & 4.81 & 1.10 & \second{3.81} & \bestnum{0.31} & \second{9.82} & \second{1.21} & 84 \\
\textbf{Lite Any Stereo} & \bestnum{3.04} & \bestnum{0.79} & \bestnum{3.87} & \bestnum{0.99} & \bestnum{3.53} & \second{0.32} & \bestnum{7.51} & \bestnum{0.94} &  \bestnum{33} \\
\hline
\hline
\rowcolor{gray!10} \multicolumn{10}{l}{\hspace{0pt}\textit{Accurate methods: Million-scale}}\\
Selective-IGEV$^{\ddag}$ \cite{wang2024selective} & 3.20 & -- & 4.50 & -- & 3.40 & -- & 7.50 & -- & 3619 \\
FoundationStereo \cite{wen2025foundationstereozeroshotstereomatching} & 2.51 & 0.67 & 2.83 & 0.86 & 0.49 & 0.14 & 1.12 & 0.37 & 12824 \\
\bottomrule
\end{tabular}
\vspace{-.6em}
\label{tab:comparison_rvc}
\end{table*}

\begin{table*}[t] \addtolength{\tabcolsep}{2pt}
\caption{Comparison of model performance on DrivingStereo weather \cite{yang2019drivingstereo}. Lower value is better for both metrics.}
\vspace{-.6em}
\centering
\small
\begin{tabular}{lccccccccccc}
\toprule
\multirow{2}{*}{Method} & \multicolumn{2}{c}{Rainy} & \multicolumn{2}{c}{Sunny} & \multicolumn{2}{c}{Foggy} & \multicolumn{2}{c}{Cloudy} & \multicolumn{2}{c}{Overall} & \multirow{2}{*}{MACs (G)}\\
 & D1 & EPE & D1 & EPE & D1 & EPE & D1 & EPE & D1 & EPE  \\
\hline
FoundationStereo \cite{wen2025foundationstereozeroshotstereomatching} & 27.01 & 3.96 & 4.31 & 1.57 & 7.67 & 1.82 & 3.85	& 1.53 & 10.71 & 2.22 & 12824 \\
\textbf{Lite Any Stereo}  & \bestnum{20.69} & \bestnum{2.61} & \bestnum{3.84} & \bestnum{1.47} & \bestnum{6.78} & \bestnum{1.64} & \bestnum{3.65} & \bestnum{1.47} & \bestnum{8.74} & \bestnum{1.80} & \bestnum{33} \\
\bottomrule
\end{tabular}
\vspace{-.6em}
\label{tab:drivingstereo-weather}
\end{table*}

\subsection{Benchmarks, Metrics, and Baselines}
\noindent \textbf{Benchmarks.} We evaluate our method on five widely-used real-world stereo datasets. {Middlebury}~\cite{middlebury} is an indoor dataset containing 15 stereo pairs with high-quality ground truth captured using structured light. In the main paper, we report results under half-resolution and non-occluded settings; full-resolution and quarter-resolution results are provided in the supplementary materials. {ETH3D}~\cite{eth3d} consists of 27 grayscale stereo pairs with laser-scanned ground truth, covering both indoor and outdoor scenes. {KITTI 2012}~\cite{kitti12} and {KITTI 2015}~\cite{kitti15} include 194 and 200 stereo pairs, respectively, captured in outdoor driving environments, with ground truth obtained via LiDAR. DrivingStereo weather \cite{yang2019drivingstereo} contains driving scene images under four different weather conditions, where each class of weather contains 500 frames. We report results at full resolution. Note that none of these datasets are used in the training process.

\noindent \textbf{Evaluation Metrics.} For all datasets, we report the average End-Point Error (EPE), which measures the mean per-pixel disparity error. For Middlebury and ETH3D, we also report the percentage of pixels with disparity errors greater than a threshold $X$ (Bad-$X$). For KITTI datasets, we report the D1 error, defined as the percentage of pixels with disparity error exceeding 3 pixels and 5\% of the ground truth disparity. 

\noindent \textbf{Baselines.} To ensure fair comparison, we re-evaluate all baseline methods under consistent settings on our local machine. This avoids discrepancies caused by differing benchmark configurations, such as occlusion masking and metric definitions (e.g., D1 vs. Bad-3.0). Unless otherwise noted, we use the official checkpoints provided in each method.

\begin{figure*}[t]
   \begin{center}
   \includegraphics[width=1\linewidth]{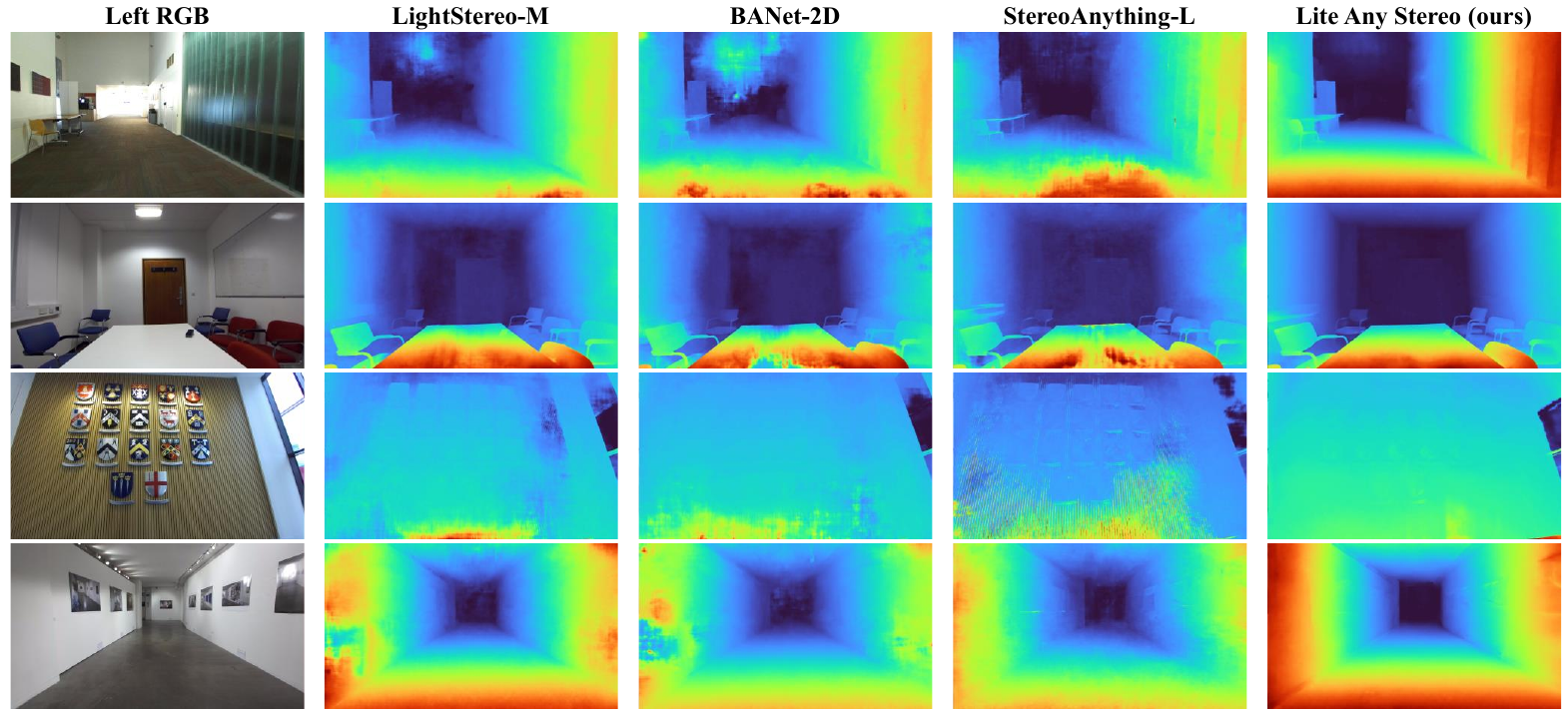}
   \end{center}
   \vspace{-1.2em}
    \caption{Qualitative comparison of zero-shot inference on in-the-wild images. Each column shows disparity predictions from different methods using the same checkpoint (Tab. \ref{tab:comparison_rvc}, second block), trained on million-scale data. The selected scenarios contain diverse real-world challenges, including reflections, complex lighting conditions, non-textured and repetitive textures. Comparison methods produce blurry boundaries, texture-copy artifacts, or fail to recover geometric details, while our method consistently preserves structural integrity, produces smooth yet accurate disparity maps, and demonstrates superior zero-shot generalization across diverse scenarios.} 
    \vspace{-1.3em}
   \label{fig:qualitative comparison}
\end{figure*}

\subsection{Implementation Details}
Lite Any Stereo is implemented using PyTorch. The model is trained for 150K, 50K, and 100K steps in stage 1, stage 2, and stage 3 with a total batch size of 176 on NVIDIA A100 GPUs. We adopt the AdamW optimizer~\cite{adam} with a one-cycle learning rate schedule, where peak learning rate is set to $2 \times 10^{-4}$. During training, input images are randomly cropped to $256 \times 512$, and then finetuned on $320 \times 736$. The maximum disparity value $\text{D}_{\text{max}}$, is set to 192, following the same configuration in previous works~\cite{guo2024lightstereochannelboostneed,xu2025banet}.

\subsection{Ablation Study} \label{sec:Ablation Study}
We investigate various design choices and training strategies for our model. Unless otherwise specified, all models are trained for 100K iterations without data augmentation on a 1.4M-image subset of synthetic datasets \cite{wen2025foundationstereozeroshotstereomatching, li2022practical, cabon2020vkitti2, fallingthings, mayer2016large}. We report D1 errors on KITTI 2012 \cite{kitti12} (K.12) and KITTI 2015 \cite{kitti15} (K.15), Bad 1.0 error on ETH3D \cite{eth3d} (E.), and Bad 2.0 error on Middlebury \cite{middlebury} (M.), as in Tab.~\ref{tab:mae_ablation}.

\noindent \textbf{Cost Aggregation Design.} In Tab.~\ref{tab:mae_ablation}(a), we first compare different aggregation choices (Fig.~\ref{fig:backbone_comparison}) and observe that the 3D–2D hybrid design is effective. We then evaluate alternative 2D layer choices in Tab.~\ref{tab:mae_ablation}(c), and find that ConvNeXt layers \cite{liu2022convnet} provide the best trade-off between accuracy and efficiency. Building upon the ConvNeXt-based 2D layers, we further evaluate the influence of 3D convolution kernel sizes (Tab.~\ref{tab:mae_ablation}(b)). Unlike the findings in \cite{wen2025foundationstereozeroshotstereomatching}, where larger kernels improved performance, our results show that standard $(3,3,3)$ configuration achieves the highest accuracy in our baseline. Finally, as shown in Tab.~\ref{tab:mae_ablation}(d), increasing the proportion of 3D blocks leads to performance degradation, particularly on the Middlebury dataset under limited MACs. Therefore, we retain only a small 3D component to preserve disparity perception while maintaining efficiency.

\noindent \textbf{Training Strategy Choices.} As shown in Tab.~\ref{tab:mae_ablation}(e) and (f), we compare different training strategies in stage \ding{173}. Our self-distillation approach yields more robust feature representations than data augmentation, leading to stronger performance. Furthermore, keeping the teacher model’s weights fixed proves more effective than using EMA or hard copy updates. Tab.~\ref{tab:mae_ablation}(g) highlights the benefits of our three-stage training scheme, which overall improves performance. The slight drop on Middlebury in stage 3 may be attributed to the limited scale of the indoor real-world data.

\begin{table*}[t] \addtolength{\tabcolsep}{2pt}
\caption{Results on {KITTI 2012}~\cite{kitti12} and {KITTI 2015}~\cite{kitti15} leaderboard. The MACs are measured for an input size of $1242 \times 375$. Our finetuned version ranks 1st on the leaderboards among all efficient methods at the time of submission.}
\vspace{-.6em}
\centering
\small
\begin{tabular}{lccccc|cccc}
\toprule
\multirow{2}{*}{Method} & \multicolumn{5}{c|}{{KITTI 2012}} & \multicolumn{3}{c}{{KITTI 2015}} & \multirow{2}{*}{MACs (G)} \\
& 3-noc & 3-all & 4-noc & 4-all & EPE-noc / all & D1-bg & D1-fg & D1-all & \\
\midrule
DispNetC~\cite{mayer2016large} & 4.11 & 4.65 & 2.77 & 3.20 & 0.9 / 1.0 & 4.32 & 4.41 & 4.34 & - \\
DeepPruner-Fast~\cite{duggal2019deeppruner} & - & - & - & - & - & 2.32 & 3.91 & 2.59 & 194 \\
AANet+~\cite{xu2020aanet} & 1.55 & 2.04 & 1.20 & 1.58 & \bestnum{0.4 / 0.5} & 1.65 & 3.96 & 2.03 & - \\
DecNet~\cite{yao2021decomposition} & - & - & - & - & - & 2.07 & 3.87 & 2.37 & - \\
BGNet+~\cite{xu2021bilateral} & 1.62 & 2.03 & 1.16 & 1.48 & 0.5 / 0.6 & 1.81 & 4.09 & 2.19 & 77\\
HITNet~\cite{tankovich2021hitnet} & 1.41 & 1.89 & 1.14 & 1.53 & \bestnum{0.4 / 0.5} & 1.74 & 3.20 & 1.98 & {47} \\
CoEx~\cite{bangunharcana2021correlate} & 1.55 & 1.93 & 1.15 & 1.42 & 0.5 / 0.5 & 1.79 & 3.82 & 2.13 & 54 \\
MobileStereoNet-2D~\cite{shamsafar2022mobilestereonet} & - & - & - & - & - & 2.49 & 4.53 & 2.83 & 127 \\
MobileStereoNet-3D~\cite{shamsafar2022mobilestereonet} & - & - & - & - & - & 2.75 & 3.87 & 2.10 & 564\\
Fast-ACVNet~\cite{xu2022acvnet} & 1.68 & 2.13 & 1.23 & 1.56 & 0.5 / 0.6 & 1.82 & 3.93 & 2.17 & 72 \\
Fast-ACVNet+~\cite{xu2023accurate} & 1.45 & 1.85 & 1.06 & 1.36 & 0.5 / 0.5 & 1.70 & 3.53 & 2.01 & 85 \\
Lite-CREStereo++ \cite{Jing_2023_ICCV} & 1.43 & 1.82 & 1.12 & 1.44 & 0.5 / 0.5 & 1.79 & 3.53 & 2.08 & 93\\
LightStereo-M \cite{guo2024lightstereochannelboostneed} & 1.56 & 1.91 & 1.10 & 1.36 & 0.5 / 0.5 & 1.81 & 3.22 & 2.04 & \bestnum{33} \\
LightStereo-L \cite{guo2024lightstereochannelboostneed} & 1.55 & 1.87 & 1.10 & 1.33 & 0.5 / 0.5 & 1.78 & \bestnum{2.64} & 1.93 & 84\\
BANet-2D \cite{xu2025banet} & 1.38 & 1.79 & 1.01 & 1.32 & 0.5 / 0.5 & {1.59} & {3.03} & {1.83} & {36} \\
BANet-3D \cite{xu2025banet} & \second{1.27} & \second{1.72} & \second{0.95} & \second{1.27} & 0.5 / 0.5 & \second{1.52} & \second{3.02} & \second{1.77} & 78 \\
\midrule
\textbf{Lite Any Stereo} & \bestnum{1.09} & \bestnum{1.49} & \bestnum{0.76} & \bestnum{1.04} & \bestnum{0.4 / 0.5} & \bestnum{1.36}	& {3.45} & \bestnum{1.71} & \bestnum{33}  \\
\bottomrule
\end{tabular}
\vspace{-1.3em}
\label{tab:kitti_benchmark}
\end{table*}

\subsection{Evaluation}
\noindent \textbf{Zero-Shot Generalization.}  Tab.~\ref{tab:comparison_rvc} presents a quantitative comparison of zero-shot generalization performance across four public datasets. In the first block of the table, we restrict our model training to stage 2 on SceneFlow to ensure a fair comparison. Under this setting, our method achieves superior performance on most metrics while requiring the fewest MACs, highlighting the effectiveness of both model design and training strategy. We then evaluate under a more realistic setting as in~\cite{wen2025foundationstereozeroshotstereomatching}, where methods are allowed to train on any available dataset except the target domains. In this setup, our approach consistently outperforms the other methods by a large margin. Our model  achieves comparable or better results than non-prior-based accurate method \cite{wang2024selective} with less than 1\% of its MACs. 
Fig.~\ref{fig:qualitative comparison} presents qualitative comparison on in-the-wild, high-resolution ($2$K) images. 
Our method generates smooth yet accurate disparity maps across different real-world scenes, whereas the other methods struggle to preserve geometric details.

\begin{table}[t] \small\addtolength{\tabcolsep}{7pt}
\caption{Performance of the proposed training strategy.}
\vspace{-.6em}
\centering
\begin{tabular}{l|cccc}
\toprule
{Method} & K.12 & K.15 & E. & M. \\
\midrule
\rowcolor{gray!10} \multicolumn{5}{l}{\hspace{0pt}\textit{LightStereo-M} \cite{guo2024lightstereochannelboostneed}}\\
Stage \ding{172}       & 4.34 & 5.27 & 6.68 & 10.29  \\
Stage \ding{173}       & 3.80 & 4.62 & 5.44 & \textbf{8.96} \\
Stage \ding{174}       & \textbf{3.35}  & \textbf{4.14}  & \textbf{4.22}  & 9.85  \\
\midrule
\rowcolor{gray!10} \multicolumn{5}{l}{\hspace{0pt}\textit{BANet-2D} \cite{xu2025banet}}\\
Stage \ding{172}       & 4.34 & 4.78 & 7.71 & 10.54  \\
Stage \ding{173}       & 3.87  & 4.80  & 4.86  & \textbf{9.54}  \\
Stage \ding{174}       & \textbf{3.28} & \textbf{4.08} & \textbf{4.05} & 10.30  \\
\bottomrule
\end{tabular}
\vspace{-1.2em}
\label{tab:universal training strategy}
\end{table}

\noindent \textbf{In-Domain Performance.} Tab.~\ref{tab:drivingstereo-weather} reports the results on the DrivingStereo weather subset. Interestingly, although our model is only trained via knowledge distillation on general datasets, it surpasses the teacher model (FoundationStereo \cite{wen2025foundationstereozeroshotstereomatching}) by a significant margin. This highlights the strong effectiveness of our proposed approach. Although in-domain performance is not the primary focus of this work, we also evaluate our model on the KITTI online leaderboard (test set) trained on \cite{kittidepth} without using its original annotations. As can be seen in Tab.~\ref{tab:kitti_benchmark}, our model achieves the highest accuracy among all published efficient methods across almost all metrics at the time of submission.

\noindent \textbf{Effectiveness of the Training Strategy.}  To verify the generality of our training strategy, we apply it to two representative efficient methods, LightStereo-M \cite{guo2024lightstereochannelboostneed} and BANet-2D \cite{xu2025banet}. In Tab.~\ref{tab:universal training strategy}, our strategy consistently improves both methods across multiple benchmarks, which shows its broad applicability to different architectures.


\noindent \textbf{Inference Time Analysis.} Since prior works often report inference times on different hardware, direct comparisons can be unreliable. To ensure a fair comparison, we benchmark all methods locally under the same experimental setup across multiple GPU platforms. As shown in Tab.~\ref{tab:inference time}, our method ranks among the fastest across different GPUs, demonstrating strong computational efficiency. These results suggest its practical potential for deployment on embedded devices and other resource-constrained platforms.

\subsection{Limitations}
While our method improves generalization, it still trails depth-prior-based approaches. A major bottleneck is the limited availability of high-quality real-world stereo data, which restricts performance gains. For instance, accuracy on Middlebury drops. Robustness under challenging cases such as transparency and reflection also can be improved.


\begin{table}[t] \small\addtolength{\tabcolsep}{1pt}
\caption{Inference time (ms) measured locally on different GPUs under the same experimental setup (KITTI resolution), with \texttt{torch.compile} enabled and CUDA synchronization applied.}
\vspace{-.6em}
\centering
\begin{tabular}{l|cccc}
\toprule
{Method}  & 4090 & A5000 & A100 & H200\\
\midrule
Lite-CREStereo++ \cite{Jing_2023_ICCV} & 14 & 30 & 20 & 16 \\
LightStereo-M \cite{guo2024lightstereochannelboostneed} & 4 &  8 & 6 & 5  \\
LightStereo-L \cite{guo2024lightstereochannelboostneed} & 7 &  16 & 10 & 7 \\
BANet-2D \cite{xu2025banet} & 4 &  8 & 7 & 5 \\
BANet-3D \cite{xu2025banet} & 6 &  12 & 10 & 7 \\
\textbf{Lite Any Stereo} & 4 &  9 & 7 & 6  \\
\bottomrule
\end{tabular}
\vspace{-1.5em}
\label{tab:inference time}
\end{table}

\section{Conclusion}
In this paper, we presented Lite Any Stereo, an efficient zero-shot stereo matching model. By integrating a compact backbone with a three-stage training strategy on million-scale data, our model generalizes well to in-the-wild scenarios while maintaining high efficiency. Extensive evaluations demonstrate state-of-the-art performance across multiple standard benchmarks, underscoring the potential of lightweight models for broad real-world deployment. Future work includes scaling real-world data and developing a model zoo with different model sizes for flexible deployment across diverse computational budgets.


{
    \small
    \bibliographystyle{ieeenat_fullname}
    \bibliography{main}
}


\end{document}